\documentclass[letterpaper]{article} 
\usepackage{aaai25}  
\usepackage{times}  
\usepackage{helvet}  
\usepackage{courier}  
\usepackage[hyphens]{url}  
\usepackage{graphicx} 
\usepackage{natbib}  
\usepackage{caption} 
\usepackage{algorithm}
\usepackage{algorithmic}

\usepackage{subcaption}
\usepackage{threeparttable}
\usepackage{booktabs}
\usepackage{enumerate}
\usepackage{multirow}
\usepackage{amsmath}
\usepackage{amssymb}
\usepackage{tabularx}

\usepackage{newfloat}
\usepackage{listings}
\DeclareCaptionStyle{ruled}{labelfont=normalfont,labelsep=colon,strut=off} 
\floatstyle{ruled}
\newfloat{listing}{tb}{lst}{}
\floatname{listing}{Listing}
\title{Scalable Autoregressive Image Generation with Mamba}
\author{
    Haopeng Li\textsuperscript{\rm}\equalcontrib, \hspace{0.2cm}
    Jinyue Yang\textsuperscript{\rm 1,2}\equalcontrib, \hspace{0.2cm}
    Kexin Wang\textsuperscript{\rm 1,2}, \hspace{0.2cm}
    Xuerui Qiu\textsuperscript{\rm 1,2}, \hspace{0.2cm}\\
    Yuhong Chou\textsuperscript{\rm 2,3}, \hspace{0.2cm}
    Xin Li\textsuperscript{\rm 4}, \hspace{0.2cm}
    Guoqi Li\textsuperscript{\rm 1,2}\thanks{Corresponding authors.}
}
\affiliations{
    \vspace{6pt}
    \textsuperscript{\rm 1}University of Chinese Academy of Sciences\hspace{0.2cm}
    \textsuperscript{\rm 2}Institute of Automation, Chinese Academy of Sciences\\
    \textsuperscript{\rm 3}The Hong Kong Polytechnic University\hspace{0.2cm}
    \textsuperscript{\rm 4}China University of Petroleum
}

\usepackage{bibentry}

\begin{document}

\twocolumn[{
    \renewcommand\twocolumn[1][]{#1}
    \maketitle
    \vspace{-0.8cm}
    \begin{center}
        \centering\captionsetup{type=figure}
        \includegraphics[width=\linewidth]{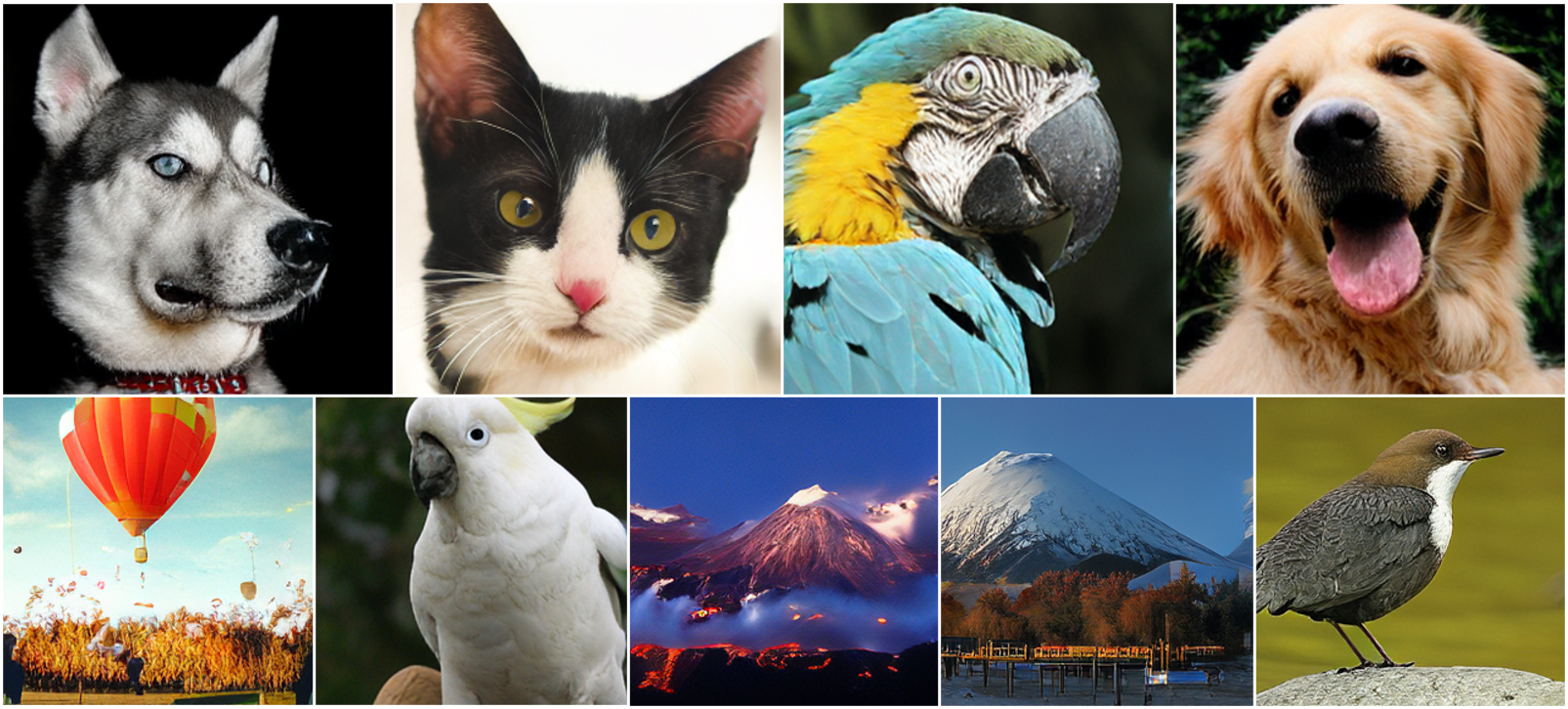}
        \captionof{figure}{\textbf{Autoregressive Image Generation with Mamba}. We show samples from our class-conditional AiM-XL model trained on ImageNet at 256\(\times\)256 resolution.}
        \label{fig:title}
    \end{center}
}]

\renewcommand{\thefootnote}{}
\footnote{\textasteriskcentered\  Equal contribution.}
\footnote{† Corresponding author.}
\renewcommand{\thefootnote}{1}

\begin{abstract}
We introduce AiM, an autoregressive (\textbf{A}R) \textbf{i}mage generative model based on \textbf{M}amba architecture. AiM employs Mamba, a novel state-space model characterized by its exceptional performance for long-sequence modeling with linear time complexity, to supplant the commonly utilized Transformers in AR image generation models, aiming to achieve both superior generation quality and enhanced inference speed. Unlike existing methods that adapt Mamba to handle two-dimensional signals via multi-directional scan, AiM directly utilizes the next-token prediction paradigm for autoregressive image generation. This approach circumvents the need for extensive modifications to enable Mamba to learn 2D spatial representations. By implementing straightforward yet strategically targeted modifications for visual generative tasks, we preserve Mamba's core structure, fully exploiting its efficient long-sequence modeling capabilities and scalability. We provide AiM models in various scales, with parameter counts ranging from 148M to 1.3B. On the ImageNet1K 256×256 benchmark, our best AiM model achieves a FID of 2.21, surpassing all existing AR models of comparable parameter counts and demonstrating significant competitiveness against diffusion models, with 2 to 10 times faster inference speed. 
Code is available at \url{https://github.com/hp-l33/AiM}

\end{abstract}

%

\begin{figure*}[t]
    \centering
    \includegraphics[width=\linewidth]{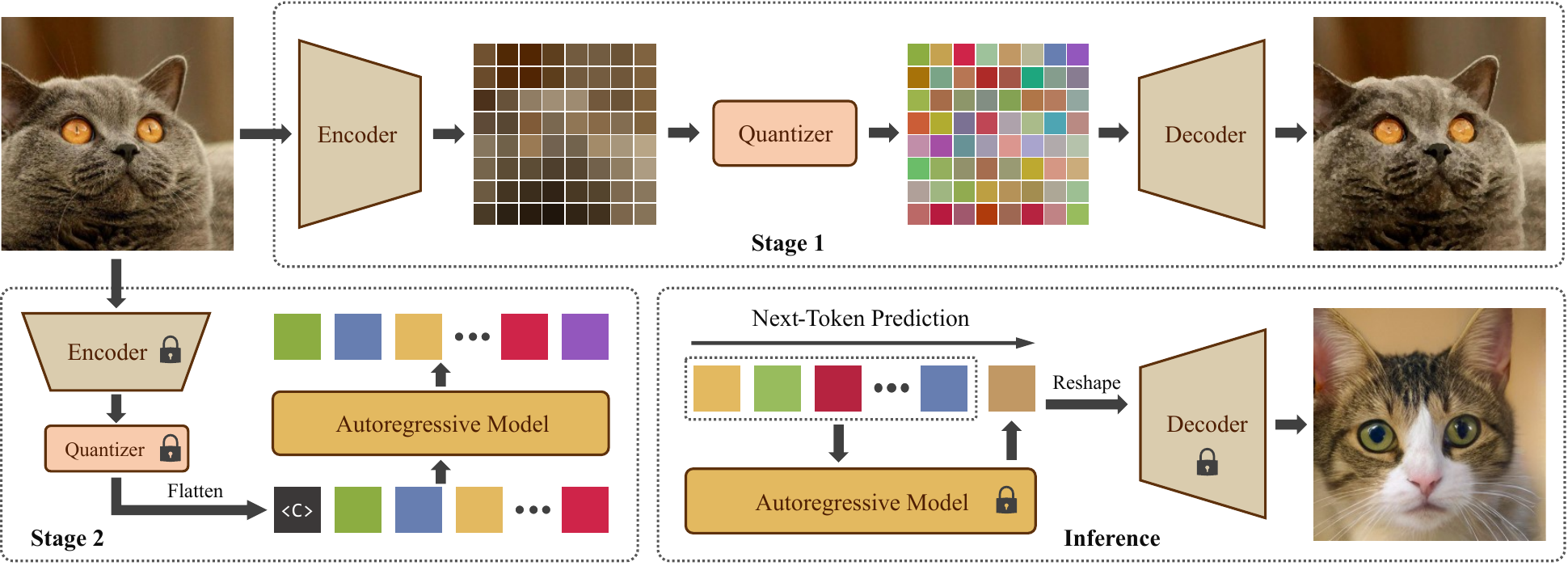}
    \caption{\textbf{AR image generation pipeline.} \textbf{Stage 1:} Training the image tokenizer (encoder and quantizer) and decoder via image reconstruction. \textbf{Stage 2:} Training the AR model through causal sequence modeling. The symbol \(\langle\text{C}\rangle\) represents the class embedding. \textbf{Inference:} Generating image tokens autoregressively by predicting the next token, which the decoder then converts into a synthesized image. \textbf{The lock icon:} Frozen weights.}
    \label{fig:pipeline}
\end{figure*}

\section{Introduction}\label{sec:intro}
In recent years, autoregressive models, particularly those based on the Transformer Decoder architecture~\cite{c:22}, have revolutionized large language models (LLMs)~\cite{van2017neural,radford2019language}. These models, which operate on the “next token prediction” paradigm, have demonstrated unprecedented performance and scalability~\cite{kaplan2020scalinglawsneurallanguage,hoffmann2022trainingcomputeoptimallargelanguage,wei2022emergentabilitieslargelanguage,henighan2020scalinglawsautoregressivegenerative}, profoundly impacting generative tasks.

Building on this success, researchers have begun exploring the capabilities of large autoregressive models for visual generation tasks. Notable models such as VQGAN~\cite{Esser_Rombach_Ommer_2021} and DALL-E~\cite{ramesh2021zeroshottexttoimagegeneration} have adapted the autoregressive approach by converting continuous images into discrete tokens and generating these tokens sequentially, achieving state-of-the-art performance at the time~\cite{yu2021vector,ramesh2021zero}. However, the emergence of diffusion models has since set new benchmarks, surpassing autoregressive models in performance.

Despite their temporary eclipse by diffusion models, the scalability of diffusion models remains limited, whereas autoregressive models offer superior scalability, making them more suitable for large-scale applications. Moreover, diffusion models follow a fundamentally different paradigm from autoregressive language models, posing significant challenges for unifying language and vision models. This ongoing challenge has motivated continued research into autoregressive visual generation models.

Recent advancements have shown promising results, with autoregressive models achieving generation quality that rivals or exceeds that of diffusion models. Key innovations include next-scale prediction~\cite{tian2024visualautoregressivemodelingscalable} techniques and the incorporation of advanced architectures like Llama~\cite{touvron2023llamaopenefficientfoundation, sun2024autoregressive}. Despite these advances, challenges remain, particularly in computational efficiency due to the high dimensionality and complexity of visual data and the quadratic computational complexity of Transformers with respect to sequence length~\cite{lee2022autoregressive,chang2022maskgit,beltagy2020longformer}.

Efforts to address these challenges have led to the exploration of linear attention mechanisms~\cite{lingle2023transformer,sun2023retentive,peng2023rwkv} as alternatives to the traditional self-attention mechanism in Transformers. One such promising model is Mamba~\cite{gu2023mamba}, a state-space model (SSM) designed for efficient sequence modeling with linear computational complexity. Mamba has demonstrated outstanding performance in language tasks and is now being applied to the visual domain~\cite{liu2024vmambavisualstatespace,zhu2024vision}. However, its potential for autoregressive image generation remains untapped.

To address this gap, we present AiM, the first autoregressive image generation model based on the Mamba architecture. AiM employs a next-token prediction paradigm with strategic enhancements tailored for the vision domain, notably the integration of a novel adaptive layer normalization method, adaLN-Group. These enhancements optimize the balance between performance and parameter count, fully leveraging Mamba's efficient sequence modeling capabilities for class-conditional image generation. 

On the ImageNet1K 256×256 benchmark~\cite{deng2009imagenet}, AiM achieves a Fréchet Inception Distance (FID) of 2.21, outperforming existing Transformer based autoregressive models of comparable scales and demonstrating significant competitiveness against diffusion models. It is noteworthy that the smallest-scale AiM model achieves a FID of 3.5 with just 148M parameters, outperforming other models that need more than twice the parameter count for similar results. Additionally, AiM offers significantly faster inference speeds compared to both Transformer based AR models and diffusion models. 
In summary, our contributions include:

1. We introduce AiM, an autoregressive image generation model based on Mamba framework, offering high-quality and efficient class-conditional image generation. To the best of our knowledge, AiM is the first of its kind.
   
2. We have adapted the architecture specifically for visual generation tasks by incorporating positional encoding and introducing a novel, more generalized adaptive layer normalization method called adaLN-Group, which optimizes the balance between performance and parameter count.
   
3. We developed AiM at varying scales and demonstrated that our approach achieves state-of-the-art performance among AR models on the ImageNet 256×256 benchmark, while also achieving fast inference speeds. These results underscore the efficiency and scalability of AiM.

\section{Related Works}\label{sec:related}
\paragraph{VQ-based AR Generative Models} 
The VQ-VAE~\cite{van2017neural} introduced a pioneering image generation approach that compresses images into a latent space and quantizes them into discrete codes by mapping continuous representations to their nearest vectors in a fixed-size codebook. These discrete codes are then modeled with a PixelCNN~\cite{van2016pixel}, predicting the probability distribution of each code given the previous ones in a raster-scan order.
This two-stage paradigm has been foundational for many subsequent works. DALL-E~\cite{ramesh2021zeroshottexttoimagegeneration} further developed this by using the Transformer to autoregressively generate  tokens. VQGAN~\cite{Esser_Rombach_Ommer_2021} enhanced the image tokenizer with adversarial and perceptual losses, achieving impressive results. 
 Recent works like VAR~\cite{tian2024visual} and LlamaGen~\cite{sun2024autoregressive} have continued this trend, demonstrating superior performance over diffusion models~\cite{nichol2021improved}. 
 
 This two-stage paradigm decouples the generation process, allowing the second stage to focus solely on sequence modeling without inductive biases on visual signals. This enables linear complexity AR models, such as Mamba, to efficiently implement autoregressive image generation without complex modifications to adapt to visual signals.

\paragraph{State Space Models} 
SSM are a class of models designed for handling long-sequence tasks, closely related to RNN~\cite{grossberg2013recurrent} models. These models utilize hidden states \( h_t \in \mathbb{R}^N \) to model sequences, enabling the capture of temporal dependencies effectively. 
Recently, a novel SSM called Mamba~\cite{gu2023mamba} has been introduced. Mamba proposed the Selective Scan mechanism, which employs technologies like kernel fusion, parallel scan and recomputation, and solves problems such as the computational load of SSMs, creating a highly scalable network backbone for various tasks. Building on this foundation, Mamba2~\cite{dao2024transformers} introduces the theoretical framework of Structured State Space Duality (SSD), demonstrating that selective SSMs essentially function as a generalized linear attention mechanism. Owing to their linear computational complexity and powerful modeling capabilities, the Mamba family represents a novel approach with the potential to replace Transformer in long-sequence modeling tasks.

\paragraph{Mamba in Visual Generation} 
Recently, there has been preliminary exploration of Mamba's applications in the visual domain. To adapt Mamba for visual signals, researchers have adopted multi-directional scan schemes. For instance, the ViM~\cite{zhu2024vision}  employs a bi-directional scan strategy, while VMamba~\cite{liu2024vmambavisualstatespace} scans input patches along four different paths. These methods employ multiple distinct SSM blocks to independently process each directional input, subsequently merging the outputs to construct the 2D representations. 
However, these multi-directional scan methods introduce additional parameters and computational costs, diminishing Mamba's speed advantage and increasing GPU memory burden. This makes it challenging to apply Mamba in visual generation tasks. To address this, Zigma~\cite{hu2024zigma} introduced the "zigzag-scan", which incorporates eight distinct scanning directions to capture 2D spatial information, with the scan process distributed across layers. Similarly, DiM~\cite{chen2023pixart} alternates between four scan directions. 
In contrast, our work uniquely adapts Mamba to autoregressive image generation models. By maximizing its long-sequence modeling capabilities and following the next-prediction paradigm, we achieve high-quality image modeling without additional scan strategy.

\begin{figure*}[t]
    \includegraphics[width=\linewidth]{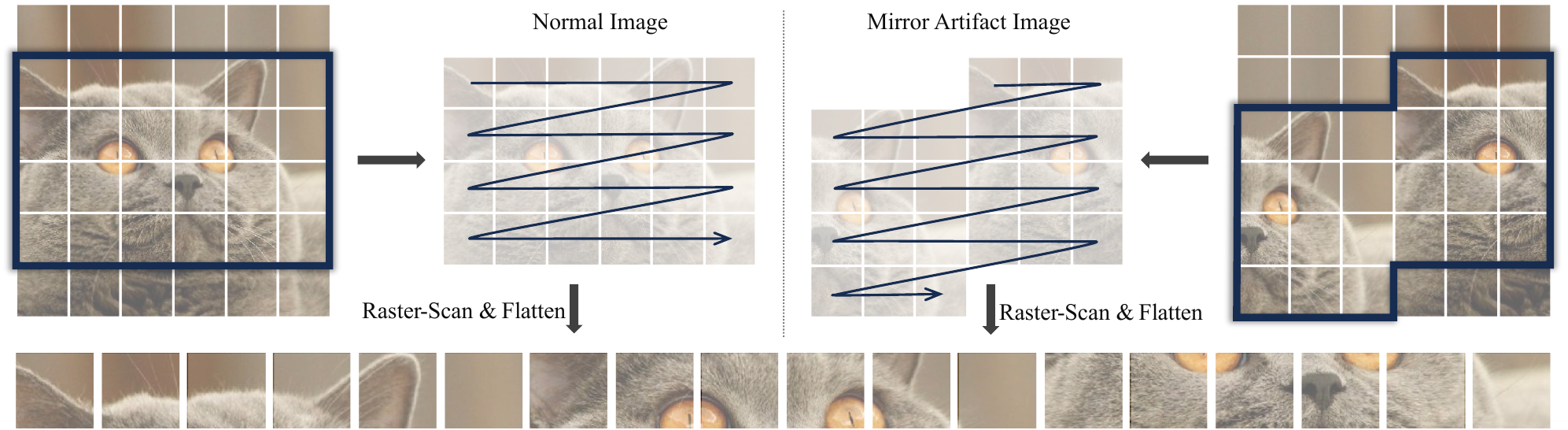}
    \caption{\textbf{The cause of mirror artifact in synthesized images}. The regions boxed in normal image and mirror mrtifact image maintain the same token sequence after flattening.}
    \label{fig:pe-reason}
\end{figure*}

\begin{figure}[t]
    \includegraphics[width=\linewidth]{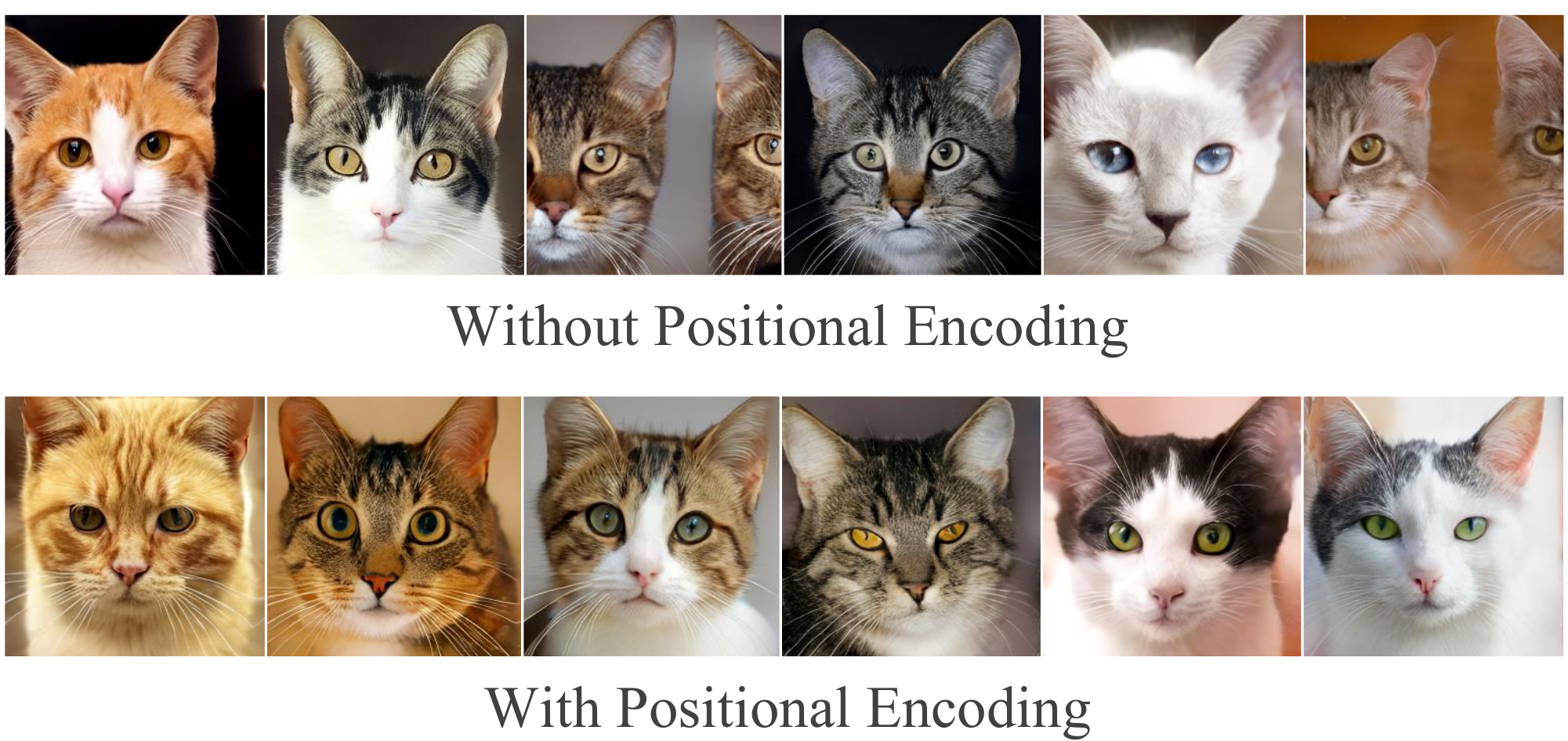}
    \caption{\textbf{The impact of positional encoding.} Without positional encoding, the model is prone to generating images with mirrored artifacts, as observed in the first row.}
    \label{fig:pe-comparison}
\end{figure}

\section{Method}\label{sec:method}
In this work, we employ the two-stage paradigm, as outlined in the previous section and depicted in Fig~\ref{fig:pipeline}. Given our primary objective to pioneer the application of Mamba in advancing autoregressive image generation, we follow the same approach as VQGAN~\cite{esser2021taming} and LDM~\cite{rombach2022high} in the first stage. The core contribution of this paper centers on the second stage.

\subsection{Preliminaries of Mamba}
The Mamba framework effectively handles sequence data for autoregressive tasks such as language modeling. It builds on state space models, which model sequences \( x(t) \in \mathbb{R} \to y(t) \in \mathbb{R} \) using hidden states \( h_t \in \mathbb{R}^N \) according to the following ordinary differential equations (ODEs) defined by parameters \( A, B, \) and \( C \):
\begin{equation}
\label{eq:h}
h'(t) = \mathbf{A}h(t) + \mathbf{B}x(t), \quad y(t) = \mathbf{C}h(t)
\end{equation}
Mamba discretizes continuous parameters using a time scale parameter \( \Delta \) through the zero-order hold (ZOH) method, transforming the ODEs for sequential data processing:
\begin{align}
\mathbf{\bar{A}} &= \exp(\mathbf{\Delta A}) \label{eq:A} \\
\mathbf{\bar{B}} &= (\mathbf{\Delta A})^{-1}(\exp(\mathbf{\Delta A}) - \mathbf{I}) \cdot \mathbf{\Delta B} \label{eq:B}
\end{align}
This allows the ODEs to be solved recurrently as follows:
\begin{equation}
\label{eq:hy}
h_t = \mathbf{\bar{A}}h_{t-1} + \mathbf{\bar{B}}x_t, \quad y_t = \mathbf{C}h_t
\end{equation}
This computing structure allows Mamba to model input sequences that perfectly match the unidirectional, next-token prediction in autoregressive modeling. By combining continuous and discrete system dynamics with dynamic parameters, Mamba effectively captures temporal dependencies and sequence patterns, making it suitable for various applications in language and vision tasks.

\subsection{Adapting for Visual Generation}
\label{subsec:adapt4cv}
Our model architecture is almost based on native Mamba, with two key improvements for adapting to the spatial properties of images and class-conditional generation.

\subsubsection{Positional Encoding} The native Mamba is not utilize positional encoding, primarily because the SSM leverages its recursive mechanism to implicitly capture positional information within sequences, which is suitable when the input data is text, given that text inherently represents a sequence progressing from left to right.
However, applying this approach to images poses challenges, as they are inherently 2-dimensional and require transformation into a  sequence, such as through raster-scan. 
In this situation, the SSM struggles to recognize ``new row'' as it can only capture sequential relationships and not accurately identify line transitions in spatial contexts. 
Such limitations can cause inaccuracies in the generated images, such as ``mirror artifact'' shown in Fig~\ref{fig:pe-reason}.
By incorporating simple absolute position encoding~\cite{dosovitskiy2020image}, we have effectively addressed the aforementioned issues, enabling the model to generate more precise and coherent images.

\begin{figure*}[t]
    \includegraphics[width=\linewidth]{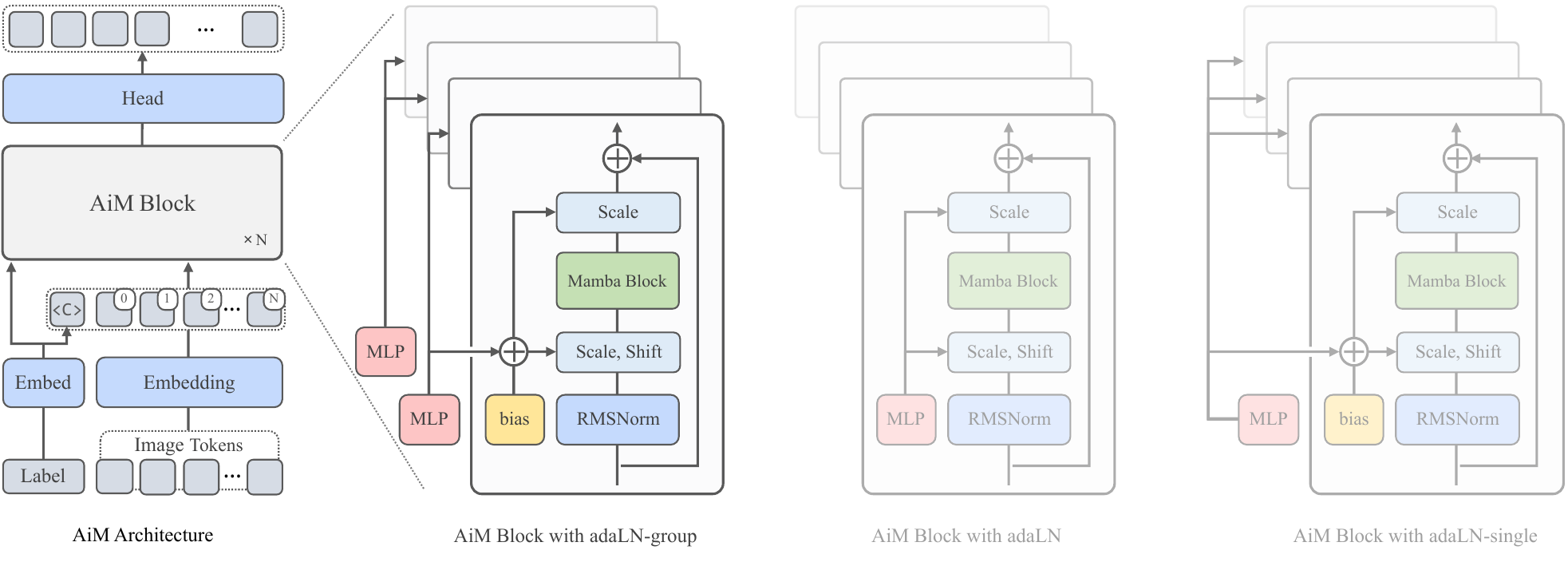}
    \caption{\textbf{Architectural details of the AiM model.} Our adaLN-group represents a more generalized form of both adaLN (when the number of groups equals the number of layers) and adaLN-single (when there is only one group)}
    \label{fig:arch}
\end{figure*}

\subsubsection{Group Adaptive Layer Normalization}
Adaptive Layer Normalization (adaLN) is a technique used to modulate data distributions based on conditional information. It has been widely adopted due to its effectiveness in various visual generation models~\cite{peebles2023scalable,perez2018film,dhariwal2021diffusion}. A mainstream variant of adaLN, proposed in DiT~\cite{peebles2023scalable} , regresses the scale parameters \(\alpha\), \(\gamma\), and the shift parameter \(\beta\) from the conditional embedding \(c\) at each layer. The normalization for the \(i\)-th layer \(F_i\) (\(i \in \{1, 2, \ldots, N\}\)) is achieved as:
\begin{equation}
[\alpha_i, \; \beta_i, \; \gamma_i]^T = \text{Swish}(c) W_i + b_i \quad \in \mathbb{R}^{3 \times d}
\end{equation}
\begin{equation}
x_i' = \gamma_i \odot F_i(\alpha_i \odot x_i + \beta_i)
\end{equation}
Where \(\text{Swish}(\cdot)\) is the Swish~\cite{ramachandran2017searchingactivationfunctions} activation function, \(d\) is embedding dimension, \(\odot\) is element-wise multiplication. While this approach improves performance, it significantly increases the parameter counts and GPU memory usage. 

To address the issue, PixArt~\cite{chen2023pixart} proposed adaLN-single, which computed the global scale and shift parameters only once and shared them across all the layers:
\begin{equation}
[\alpha, \; \beta, \; \gamma]^T = \text{Swish}(c) W + b \quad \in \mathbb{R}^{3 \times d}
\label{eq:global_param}
\end{equation}
Within each layer, the global parameters are summed with layer-specific learnable parameters to yield the final parameters used for modulation. These layer-specific parameters can be merged into the bias terms in the eq.\ref{eq:global_param} as \(b_i\):
\begin{equation}
[\alpha_i, \; \beta_i, \; \gamma_i]^T = \text{Swish}(c) W + b_i \quad \in \mathbb{R}^{3 \times d}
\end{equation}

Although adaLN-single reduces the parameter counts, it incurs a performance penalty~\cite{chen2023pixart}. To strike a better balance between parameter count and performance, we propose a more general form called \textbf{adaLN-group}. This method partitions the layers into \(G\) groups, where each group shares the local parameters regressed by a group-specific nonlinear module while each layer within the group also has layer-specific learnable parameters. For the \(i\)-th layer in the \(j\)-th group (\(j \in \{1, 2, \ldots, G\}\)):
\begin{equation}
[\alpha_i, \; \beta_i, \; \gamma_i]^T = \text{Swish}(c) W_j + b_i \quad \in \mathbb{R}^{3 \times d}
\end{equation}
Notably, when \(G=1\), adaLN-group is equivalent to adaLN-single; when \(G=N\), adaLN-group behaves identically to vanilla adaLN. 
This structure maintains a balance between parameter counts and performance by allowing groups of layers to share certain parameters while retaining individual biases. Consequently, it optimizes memory usage without significantly compromising performance.

We found that setting the number of groups to 4 achieves an optimal balance between model parameters and performance in our experiments. For a detailed discussion refer to the experiments section.

\subsection{Image Generation by Autoregressive Models}
Autoregressive image generation typically follows the next-token prediction paradigm. The key distinction in conditional generation is the inclusion of additional modality-specific information, such as class labels or text. This paper focuses exclusively on class-conditional generation.

\subsubsection{Class-conditional image generation} 
The process begins by embedding the class labels and concatenating them to the head of the image token embedding sequence. These embedded class labels simultaneously undergo a nonlinear transformation to obtain the scale and shift parameters used for adaLN. The model is trained to predict the next token in the sequence given the previous tokens. During training, the input tokens are fed into the model, which predicts the probability distribution of the subsequent token. The loss is calculated based on the discrepancy between the model's predictions and the actual target tokens, which are the input tokens shifted by one position. 
Formally, if \( q_i \) represents the \( i \)-th token and \( q_{<i} \) denotes all preceding tokens, the model predicts \( P(q_i \mid q_{<i}, c) \), where \( c \) is the class embedding. The optimization objective is to minimize the negative log-likelihood:
\begin{equation}
\label{eq:loss}
\mathcal{L} = - \sum_{i=1}^{N} \log P(q_i \mid q_{<i}, c)
\end{equation}
where \( N \) is the total number of tokens. This approach ensures that the model effectively learns to predict each token in the sequence based on the previous tokens and class label.

\subsubsection{Classifier-free guidance} In our approach, we also incorporate classifier-free guidance~\cite{dhariwal2021diffusionmodelsbeatgans} to enhance generation quality. This technique involves training the model both conditionally, with class labels, and unconditionally, without class labels. During inference, we interpolate between the unconditional model \( P(q_i \mid q_{<i}) \) and the class-conditional model \( P(q_i \mid q_{<i}, c) \). This interpolation is controlled by a guidance scale \( w \), and the resulting probability is given by:
\begin{equation}
\label{eq:cfg}
P_{\text{guide}}(q_i \mid q_{<i}, c) = P(q_i \mid q_{<i}) \cdot (1 - w) + P(q_i \mid q_{<i}, c) \cdot w
\end{equation}
This technique allows the model to adjust the influence of class labels dynamically, leading to more diverse and high-quality outputs.

\section{Experiments}
We conducted experiments on the ImageNet1K benchmark to evaluate the architectural design, performance, scalability and inference efficiency of the AiM model.
\subsection{Experimental Setup}
\subsubsection{Implementation details}
We provide AiM in four scales. Detailed configurations for each scale are provided in Tab~\ref{tab:training_config}. Unless stated otherwise, all models in the following sections 
utilize the same group setup as in Tab~\ref{tab:training_config}. Our image tokenizer is configured with a downsampling factor of 16 and is initialized with the pre-trained weights from LlamaGen.
\subsubsection{Training setup} 
We trained class-conditional AiM models on the ImageNet1K 256×256 dataset using 80GB A100 GPUs. Each image was tokenized into 256 tokens. The training process employed the AdamW optimizer with $(\beta_{1}, \beta_{2})=(0.9, 0.95)$ and a weight decay rate of 0.05. The learning rate was set to 1e-4 per 256 batch size, with the training epochs varying between 300 and 350 depending on model scale. 
A dropout rate of 0.1 was specifically applied to the class embeddings to facilitate classifier-free guidance.

\begin{table}[ht]
\centering
\begin{tabularx}{\columnwidth}{lccccc}
\toprule
Model & Params. & Layers & Dims. & Groups & Epoch\\
\midrule
AiM-B & 148M & 24 & 768  & 24 & 300\\
AiM-L & 350M & 48 & 1024 & 4 & 300\\
AiM-XL& 763M & 48 & 1536 & 4 & 350\\
AiM-1B& 1.3B & 48 & 2048 & 4 & 350\\
\bottomrule
\end{tabularx}
\caption{\textbf{Architectural design and training configuration of different models}}
\label{tab:training_config}
\end{table}

\begin{table*}[ht]
\centering
    \begin{tabularx}{0.99\linewidth}{l|lc|cc|ccc}
    \toprule
        Type                 & Model    &Params. & FID\(\downarrow\) & IS\(\uparrow\)    & Precision\(\uparrow\) & Recall\(\uparrow\) \\ 
        \midrule
        \multirow{3}{*}{GAN} & BigGAN~\cite{brock2018large}     & 112M & 6.95 & 224.5 & 0.89 & 0.38 \\
                             & GigaGAN~\cite{kang2023scaling}   & 569M & 3.45 & 225.5 & 0.84 & 0.61 \\ 
                             & StyleGanXL~\cite{sauer2022styleganxlscalingstyleganlarge}  & 166M & 2.30 & 265.1 & 0.78 & 0.53 \\
        \midrule
        \multirow{4}{*}{Diffusion} & ADM~\cite{dhariwal2021diffusion}   & 554M & 10.94 & 101.0 & 0.69 & 0.63  \\
                                   & LDM-4~\cite{rombach2022high}       & 400M &  3.60 & 247.7 & -    & -     \\
                                   & DiT-L/2~\cite{peebles2023scalable} & 458M &  5.02 & 167.2 & 0.75 & 0.57  \\
                                   & DiT-XL/2                           & 675M &  2.27 & 278.2 & 0.83 & 0.57  \\
        \midrule
        \multirow{2}{*}{Mask.}  & MaskGIT~\cite{chang2022maskgit}       & 227M  &  6.18 & 182.1 & 0.8   & 0.51  \\
                                & MaskGIT-re                            & 227M  &  4.02 & 355.6 & -     & -     \\
        \midrule
        \multirow{7}{*}{AR (Transformer)}
                            & VQGAN~\cite{esser2021taming}              & 227M & 18.65 &  80.4 & 0.78 & 0.26  \\
                            & VQGAN                                     & 1.4B & 15.78 &  74.3 & -    & -     \\
                            & VQGAN-re                                  & 1.4B &  5.20 & 280.3 & -    & -     \\
                            & ViT-VQGAN~\cite{yu2021vector}             & 1.7B &  4.17 & 175.1 & -    & -     \\
                            & ViT-VQGAN-re                              & 1.7B &  3.04 & 227.4 & -    & -     \\
                            & RQTran.~\cite{lee2022autoregressive}      & 3.8B &  7.55 & 134.0 & -    & -     \\
                            & RQTran.-re                                & 3.8B &  3.80 & 323.7 & -    & -     \\
        \midrule
        \multirow{4}{*}{VAR} 
                             & VAR-d16~\cite{tian2024visual}    & 310M & 3.30 & 274.4 & 0.84 & 0.51 \\
                             & VAR-d20                          & 600M & 2.57 & 302.6 & 0.83 & 0.56 \\
                             & VAR-d24                          & 1.0B & 2.09 & 312.9 & 0.82 & 0.59 \\
                             & VAR-d30                          & 2.0B & 1.97 & 334.7 & 0.81 & 0.61 \\
        \midrule
        \multirow{6}{*}{AR (Transformer)}
                            & LlamaGen-B   ~\cite{sun2024autoregressive} & 111M  & 5.46 & 193.6 & 0.83 & 0.45 \\
                            & LlamaGen-L                                 & 343M  & 3.81 & 248.3 & 0.83 & 0.52 \\
                            & LlamaGen-L*                                & 343M  & 3.07 & 256.1 & 0.83 & 0.52 \\
                            & LlamaGen-XL*                               & 775M  & 2.62 & 244.1 & 0.80 & 0.57 \\
                            & LlamaGen-XXL*                              & 1.4B  & 2.34 & 253.9 & 0.80 & 0.59 \\
                            & LlamaGen-3B*                               & 3.1B  & 2.18 & 263.3 & 0.81 & 0.58 \\
        \midrule
        \multirow{4}{*}{AR (Mamba)} 
                            & AiM-B     & 148M & 3.52 & 250.1 & 0.83 & 0.52 \\
                            & AiM-L     & 350M & 2.83 & 244.6 & 0.82 & 0.55 \\
                            & AiM-XL  & 763M & 2.56 & 257.2 & 0.82 & 0.57 \\
                            & AiM-1B  & 1.3B & 2.21 & 256.0 & 0.82 & 0.55 \\
    \bottomrule
    \end{tabularx}
\caption{\textbf{Model comparisons on class-conditional ImageNet 256×256 benchmark}. ``\(\downarrow\)'' or ``\(\uparrow\)'' indicate lower or higher values are better. ``-re'': rejection sampling. ``*'': the generated images are 384×384 and are resized to 256×256 for evaluation}
\label{tab:main-result}
\end{table*}

\subsubsection{Evaluation metrics} 
We used the Fréchet Inception Distance (FID)~\cite{Heusel_Ramsauer_Unterthiner_Nessler_Hochreiter_2017} as the main metric, and also took the Inception Score (IS)~\cite{salimans2016improved}, precision and recall as secondary metrics. Our baseline results were all cited from the original paper for a fair comparison.

\subsection{The Analysis of Scalability}
\begin{figure*}[t]
    \includegraphics[width=\linewidth]{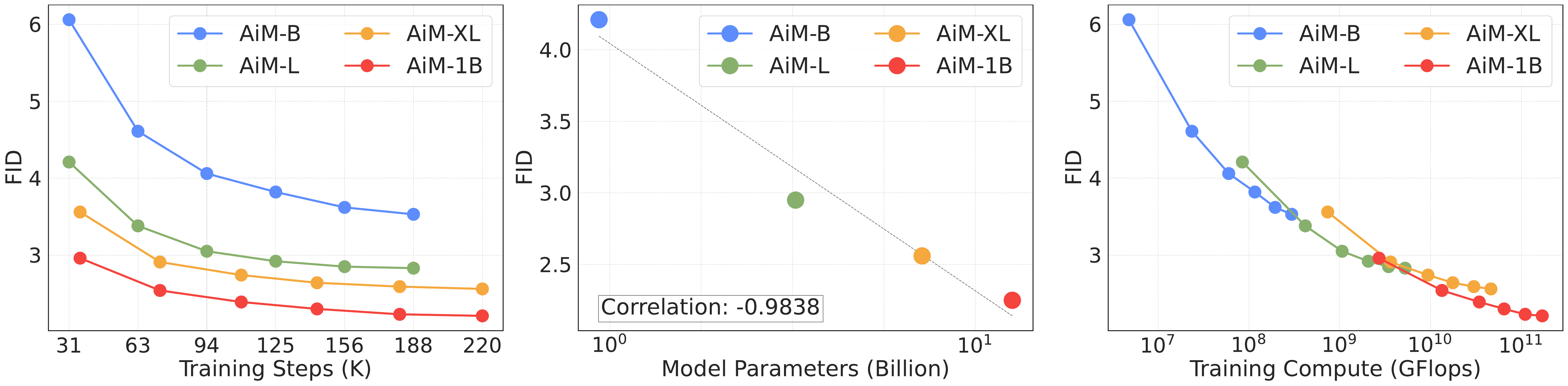}
    \caption{\textbf{AiM exhibits scalability.} \textbf{Left:} Scaling the AiM improves FID. \textbf{Center:} Model parameters strongly correlated with FID. \textbf{Right:} Larger models use large compute more efficiently.}
    \label{fig:FID Trends}
\end{figure*}

We study the scalability of AiM by varying the model parameters and the amount of training compute, assessing image quality using FID. The results are shown in Fig~\ref{fig:FID Trends}. 
FID decrease with additional training steps across all models. 
A strong correlation coefficient near -0.9838 between FID and model parameters provides solid evidence that larger models significantly improve the quality of generated images. 
These results confirm AiM's scalability, demonstrating that larger models and longer training each enhance image quality, emphasizing the need for investment in these areas for better performance. Given the constraints of the ImageNet1K, we refrained from scaling the model size to 2B or larger.

\subsection{Comparisons with Other Methods.}
We compared our models with existing generative approaches, including GANs, diffusion models, masked generative models and Transformer-based AR models across various scales, as indicated in Tab~\ref{tab:main-result}. Our AiM has achieved state-of-the-art performance in AR models and demonstrates competitive results compared to diffusion models. Samples are displayed in Fig~\ref{fig:title}.

\subsection{Ablation Study}
\subsubsection{Effect of group count in adaLN-group}
\begin{figure}[t]
    \includegraphics[width=\columnwidth]{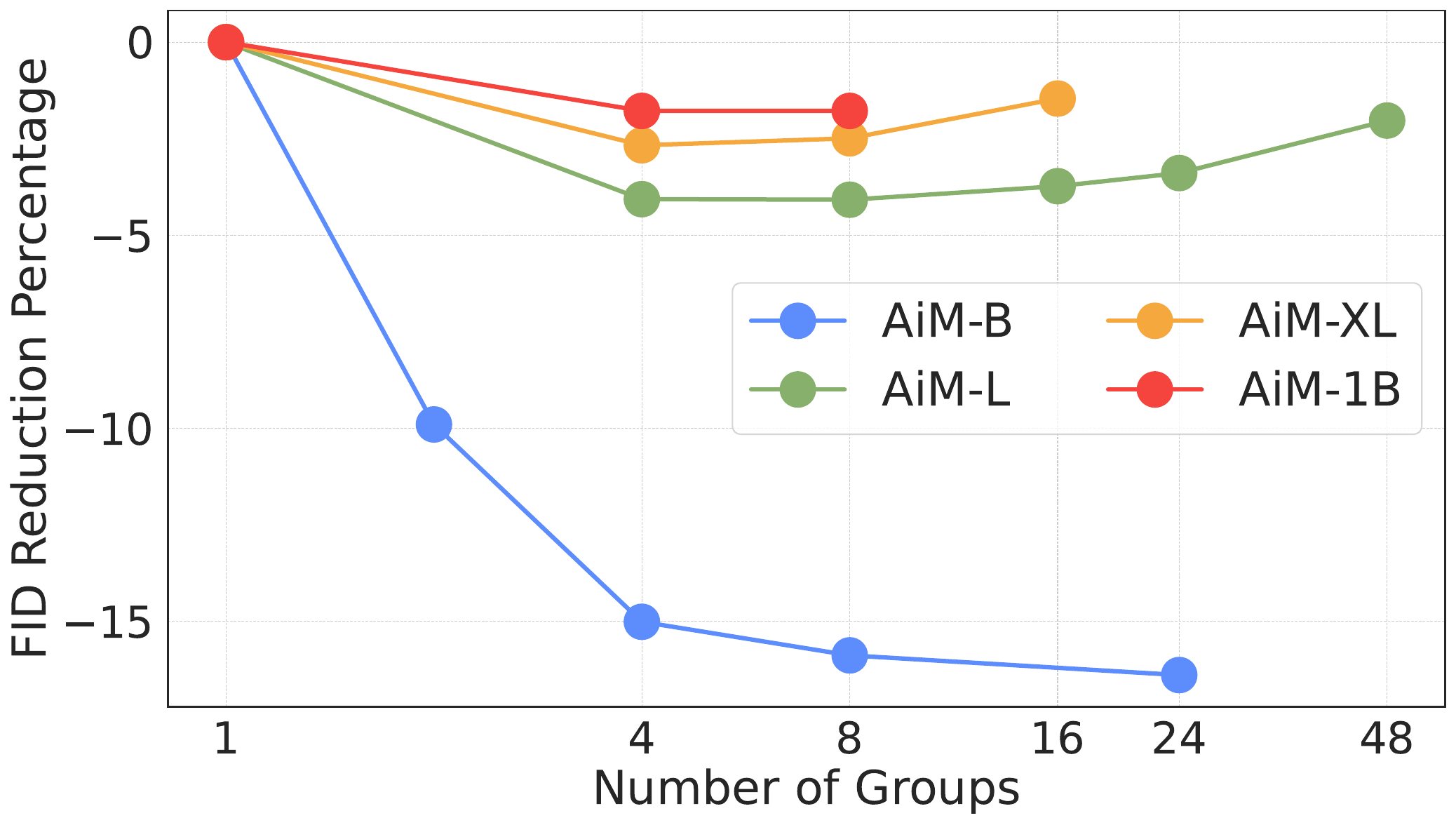}
    \caption{\textbf{Impact of group count}. A trade-off between parameter count and performance was achieved with 4 groups.}
    \label{fig:group}
\end{figure}

We first evaluated the effect of adaLN and adaLN-single on model parameter count and performance across two model scales, as detailed in Tab~\ref{table:adaln_comparison}. As the hidden size increases, the parameter count growth introduced by adaLN exhibits a non-linear relationship with performance gains, indicating redundancy. This finding highlights the need to balance parameter count and performance, motivating our exploration of adaLN-group. We further investigated the impact of group count in adaLN-group on parameter count and performance, as illustrated in Fig~\ref{fig:group}. With 4 groups, adaLN-group achieves comparable or superior performance to adaLN across model scales, confirming that excessive parameters in adaLN not only add redundancy but also complicate training.

\begin{table}[ht]
\centering
\begin{tabularx}{0.98\columnwidth}{Xl|cc|r}
\toprule
                       &                    & adaLN-single & adaLN & \(\Delta\)  \\
\midrule
\multirow{2}{*}{AiM-B} & Params. & 93M   & 134M  & \(+\)44.1\%     \\
                       & FID \(\downarrow\) & 4.21  & 3.52  & \(-\)16.4\%   \\
\midrule
\multirow{2}{*}{AiM-L} & Params. & 322M  & 470M  & \(+\)46.0\%     \\
                       & FID \(\downarrow\) & 2.95  & 2.89  & \(-\)2.0\%      \\
\bottomrule
\end{tabularx}
\caption{\textbf{Impact of adaLN-single and adaLN.} ``Params'' refers to non-embedding parameters. FID reduction shows a non-linear correlation with the growth in parameter count.}
\label{table:adaln_comparison}
\end{table}

\subsubsection{Effect of architectural enhancements} 
To validate the effectiveness of the enhanced method proposed in the previous section, we conducted an ablation study on the AiM-L model by adding these components. The CFG (default factor set to 2) significantly impacts FID, while PE has little effect on FID but a noticeable impact on visual perception. The inclusion of adaLN also significantly affects FID. More detailed experimental results can be found in the Appendix.

\begin{table}[ht]
\centering
\begin{tabularx}{0.98\columnwidth}{l|cccc|c}
\toprule
Description & Params. & adaLN & PE & CFG & FID\(\downarrow\) \\
\midrule
\multirow{4}{*}{AiM-L}  & 337M & $\times$ & $\times$ & $\times$ & 7.72 \\
                        & 337M & $\times$ & $\times$ & $\checkmark$& 3.41 \\
                        & 338M & $\times$ & $\checkmark$& $\checkmark$ & 3.34 \\
                        & 350M & $\checkmark$ & $\checkmark$& $\checkmark$ & 2.83 \\
\midrule
+ Scaling Up & 1.3B & $\checkmark$ & $\checkmark$ & $\checkmark$ & 2.21 \\
\bottomrule
\end{tabularx}
\caption{\label{tab:adaLN and PE}\textbf{Ablation study.} For simplicity, adaLN refers to the previously mentioned adaLN-group with 4 groups.}
\end{table}

\subsection{Inference Efficiency} 
\begin{figure}[t]
    \includegraphics[width=\columnwidth]{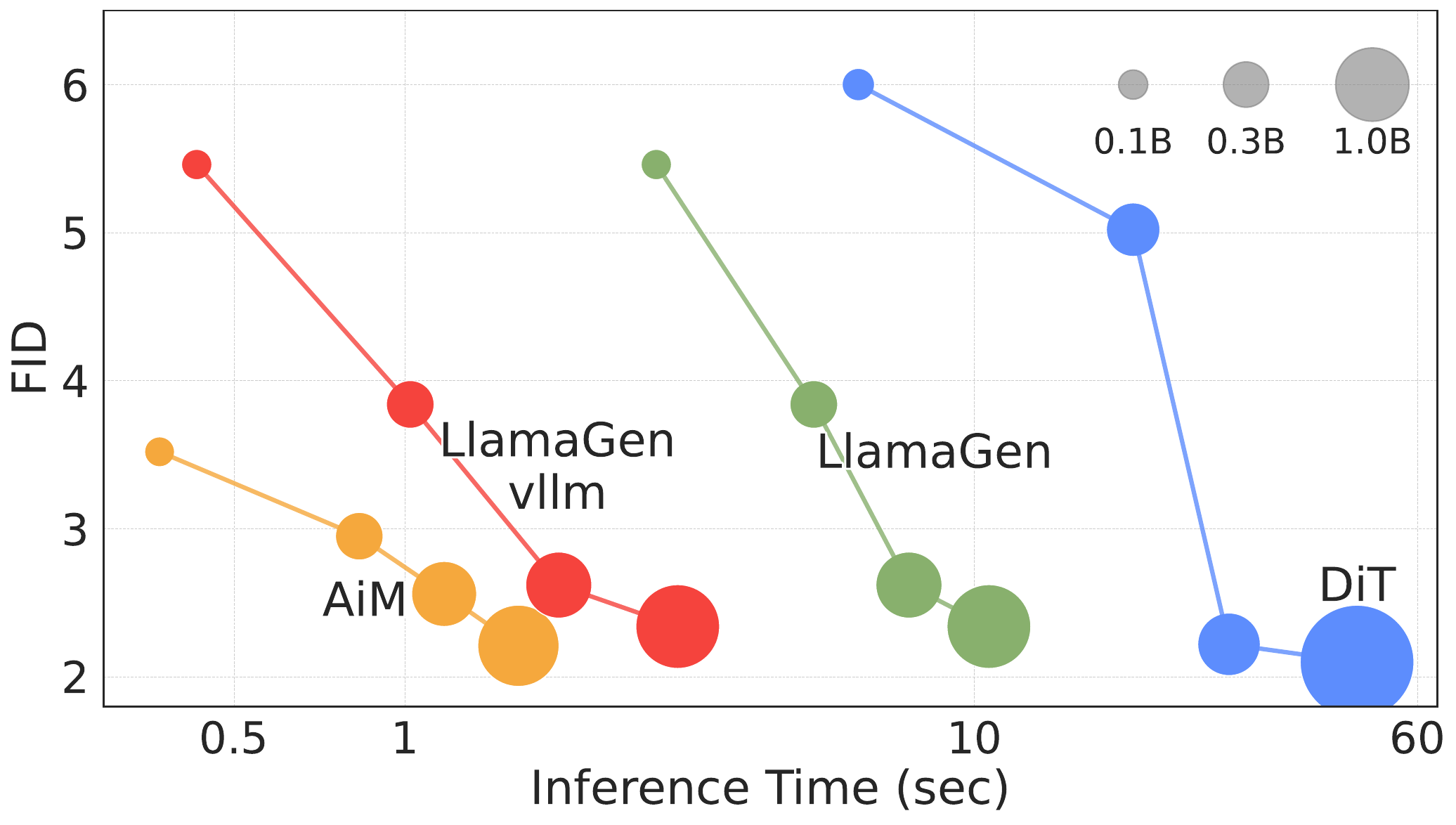}
    \caption{\textbf{Inference time on ImageNet1K 256\(\times\)256 benchmark.} Result with a batch size of 16 on the A100 GPU.}
    \label{fig:speed}
\end{figure}

We compared the inference speed of the AiM model with different models, as shown in Fig~\ref{fig:speed}. AiM demonstrates a significant advantage in inference speed. Among them, the Transformer-based models accelerate by default using Flash-Attention~\cite{dao2022flashattentionfastmemoryefficientexact} and KV Cache (only for AR models). 

\section{Conclusion}
We explore the significant potential of Mamba in visual tasks, providing insights for adapting it to visual generation without additional multi-directional scans. AiM's effectiveness and efficiency underscore its scalability and broad application potential in AR visual modeling.
However, our work has limitations: (1) We focus on class-conditional generation without exploring text-to-image generation. (2) More efficient autoregressive methods deserve further exploration. These will be addressed in our future works.

\bigskip

\bibliography{aaai25}

\bigskip

\end{document}